\documentclass[10pt,twocolumn,letterpaper]{article}

\usepackage[pagenumbers]{cvpr} 

\usepackage{graphicx}
\usepackage{amsmath}
\usepackage{amssymb}
\usepackage{booktabs}
\usepackage[T1]{fontenc} 
\usepackage{algorithm}
\usepackage{algorithmic}
\usepackage{tabularx}
\usepackage{multirow}
\usepackage{array}
\usepackage[pagebackref,breaklinks,colorlinks]{hyperref}

\usepackage[capitalize]{cleveref}
\crefname{section}{Sec.}{Secs.}
\Crefname{section}{Section}{Sections}
\Crefname{table}{Table}{Tables}
\crefname{table}{Tab.}{Tabs.}

\def\cvprPaperID{*****} 
\def\confName{CVPR}
\def\confYear{2022}

\newcolumntype{C}[1]{>{\centering\arraybackslash}p{#1}}
\newcolumntype{L}[1]{>{\arraybackslash}p{#1}}
\begin{document}

\title{Interactive Data Synthesis for Systematic Vision

Adaptation via LLMs-AIGCs Collaboration}

\author{Qifan Yu$^1$\footnotemark[1] \quad\quad Juncheng Li$^1$\footnotemark[1]  \quad\quad Wentao Ye$^1$  \quad\quad
Siliang Tang$^1$ \quad\quad Yueting Zhuang$^1$\\ $^1$Zhejiang University\\ {\tt\small \{yuqifan, junchengli, 22121058, siliang, yzhuang\}@zju.edu.cn} }

\maketitle

\renewcommand{\thefootnote}{\fnsymbol{footnote}} 
\footnotetext[1]{Equal Contribution.} 
\renewcommand{\thefootnote}{\arabic{footnote}} 

\begin{abstract}
Recent text-to-image generation models have shown promising results in generating high-fidelity photo-realistic images. In parallel, the problem of data scarcity has brought a growing interest in employing AIGC technology for high-quality data expansion. 
However, this paradigm requires well-designed prompt engineering that cost-less data expansion and labeling remain under-explored.
Inspired by LLM's powerful capability in task guidance, 
we propose a new paradigm of annotated data expansion named as \textbf{ChatGenImage}. The core idea behind it is to leverage the complementary strengths of diverse models to establish a highly effective and user-friendly pipeline for interactive data augmentation. 
In this work, we extensively study how LLMs communicate with AIGC model to achieve more controllable image generation and make the first attempt to collaborate them for automatic data augmentation for a variety of downstream tasks. Finally, we present fascinating results obtained from our ChatGenImage framework and demonstrate the powerful potential of our synthetic data for systematic vision adaptation.
Our codes are available at \url{https://github.com/Yuqifan1117/Labal-Anything-Pipeline}.
\end{abstract}

\section{Introduction}
In the past decade, deep learning techniques have demonstrated promising performance across diverse tasks, owing to the availability of large-scale annotated data~\cite{he2016deep, krizhevsky2017imagenet, girshick2015fast}.
However, it is time-consuming and expensive to manually collect a large-scale annotated dataset containing every possible domain for robust training. Besides, the problem of cross-domain and long-tail distributions within existing datasets have a detrimental effect on the performance and robustness of vision models, thereby impeding their generalization ability to novel categories or unseen domains. This promotes us to explore a less labor-intensive way to harvest labeled data containing multiple domains in one step for robust vision tasks.

One effective strategy to improve generalization and robustness is to enlarge the scale of training data by intricate augmentations~\cite{hendrycks2021many}.
There are several GAN-based models~\cite{choi2019self, jahanian2021generative} generating images for vision tasks, but their applicability remains constrained by their narrow focus on specific settings or small scales. Recently, AIGC models\cite{ramesh2022hierarchical,rombach2022high,saharia2022photorealistic} have emerged as promising candidates for generating high-quality synthetic data, with the ability to address the limitations of the existing dataset. There are several early attempts at exploring synthetic data from generative models for data augmentation~\cite{he2022synthetic, lin2023explore, bansal2023leaving, zhang2021datasetgan}.
Albeit promising, early works usually produce simple scenarios or object-centric images only by global constraints~(\textit{i.e.}, ``airplane" or ``a white airplane hovering over a beach and a city".), which limits downstream models' perception of intricate scenes and fine-grained attributes. Additionally, these methods concentrate on generating images under typical scenarios~(\textit{e.g.}, daylight, field), while neglecting less common but predictable circumstances~(\textit{e.g.}, snow, forest, night). This limitation may impede the ability of deep learning models to generalize when deployed in real-world environments that exhibit unseen test distributions.

In this paper, we present a novel approach named ChatGenImage that facilitates more controllabel data augmentation. ChatGenImage harnesses the collaborative power of the LLM and AIGC models, enabling iterative communication between them in a cost-effective and controllable manner. This automatically iterative process facilitates the generation of high-quality synthetic images depicting complex scenes and diverse domains, along with fine-grained annotations.
Our fundamental intuition is that large language models have remarkable capabilities to perform new tasks in a zero-shot manner when presented with well-crafted instruction prompts\cite{wei2021finetuned, wei2022chain, wu2023visual, gupta2022visual}. We discover that these LLMs like ChatGPT possess the capability to autonomously navigate image editing processes. By strategically designing appropriate prompts, LLMs can leverage the inherent knowledge within the system and effectively guide the AIGC models to produce highly controllable and intricate images.
While ChatGPT contains diverse world knowledge for simulating the human brain's efficient processing, it is non-trival to elicit this knowledge from it for data augmentation with automatic labeling because ChatGPT is a pure language model that lacks the ability to visually perceive any information. We explore this issue in the context of generative data augmentation, showing that language can act as a bridge connecting LLMs and AIGC models, producing elaborate images for downstream tasks by globally controllable prompts and iteratively local editing instructions.

To this end, we demonstrate three key findings. First, we find that the LLM such as \textit{ChatGPT} contains a wealth of conceptual knowledge and can imagine vivid descriptions even with only one label word~(\textit{e.g. A \textcolor{red}{dog} playing in a lush green park, with a frisbee in its mouth. The dog should have \textcolor{blue}{a shiny coat of fur}.})~\cite{touvron2023llama, bubeck2023sparks}. We further obverse that the existing AIGC models can only generate simple image with few objects and backgrounds, which are not diverse for domain generalization~\cite{kumar2022fine}. Thus, we establish the iterative pipeline to repair missing details and refine generated images with the help of label foundation toolkits and local editing prompts. Finally, we demonstrate our method flow to produce large amounts of high-quality synthetic data with fine-grained labels in a scalable manner for data augmentation in data scarcity scenarios.
\begin{figure*}[ht]

\centering
   \includegraphics[width=1.0\linewidth]{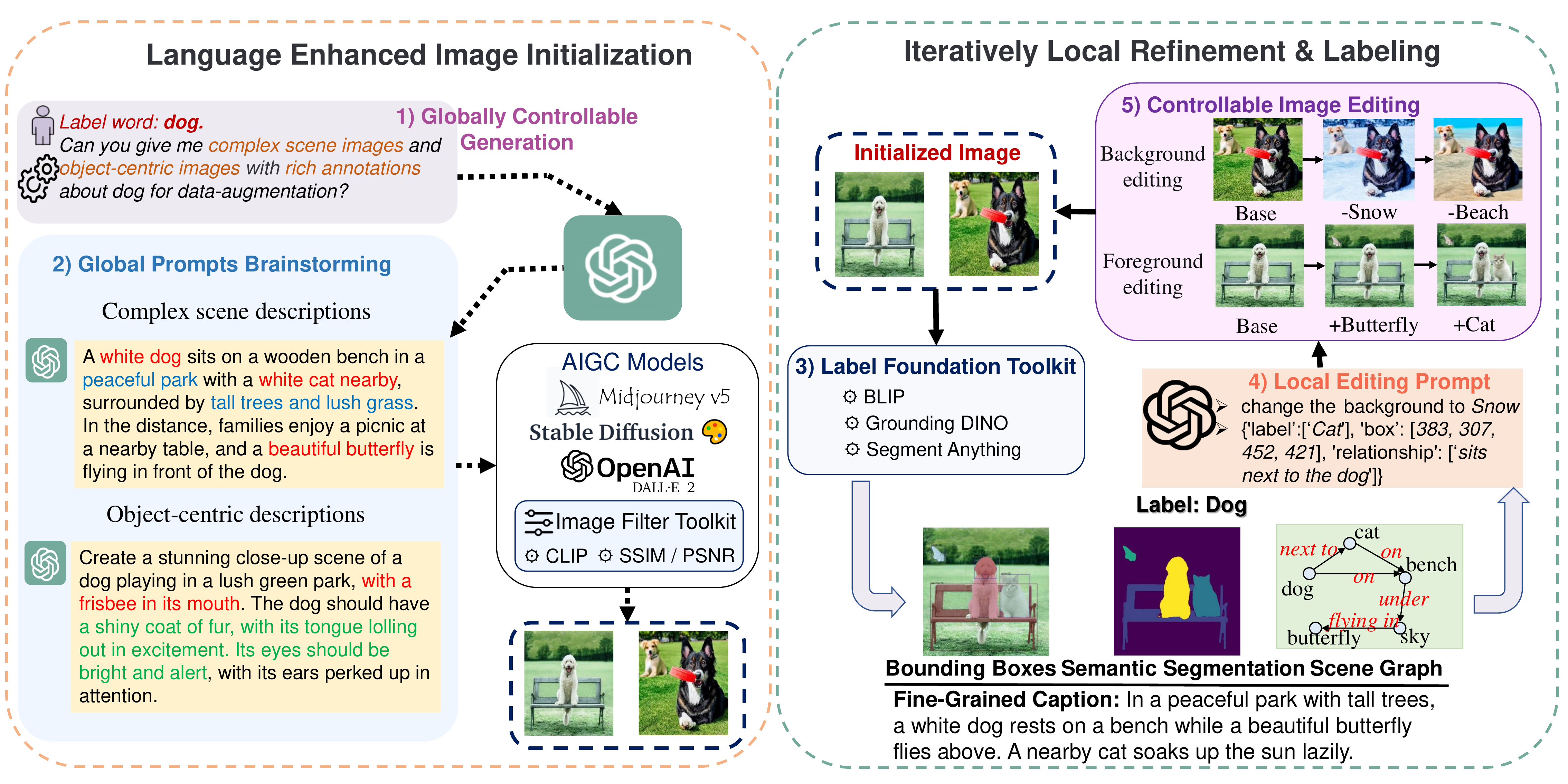}
    
   \caption{Language as a bridge for LLMs~(\textit{e.g. ChatGPT}) and AIGC models~(\textit{e.g. Stable Diffusion}) can iteratively control image generation and automatic labeling. The LLM first generates global prompts to guide AIGC models in generating initialization images, then iteratively refines them using automatically generated fine-grained annotations as local constraint prompts to produce diverse and complex scenes.}
   \label{architecturel}
\end{figure*}

\section{Related Work}
\subsection{Large Language Models} 
In recent years, the field of natural language processing has been revolutionized by the emergence of large language models~(LLMs), exemplified by models such as PaLM~\cite{chowdhery2022palm}, ChatGPT, and LLaMa~\cite{touvron2023llama}. Moreover, the remarkable performance of LLMs in zero-shot and few-shot generalization has sparked a growing trend in utilizing autoregressive language models for vision-language tasks~\cite{li2022fine, radford2021learning}. However, the generalization of LLMs does not translate well to visual tasks\cite{kumar2022fine, dunlap2022using}. Unlike previous works, we utilize LLMs to enrich training data for fine-tuning in downstream tasks instead of directly transferring by contrastive learning.
\subsection{Text-to-Image Diffusion Models}
Recently, diffusion models have become a promising generative modeling framework, achieving state-of-the-art performance on image generation tasks~\cite{ramesh2022hierarchical,rombach2022high,saharia2022photorealistic}. GLIDE~\cite{nichol2021glide} studies diffusion models for the text-conditional image synthesis by classifier-free guidance strategies. InstructPix2Pix~\cite{brooks2022instructpix2pix} proposes a effective framework to edit images with human instructions, which opens up new opportunities for controllable image creation by user-written instructions. However, existing SOTA text-to-image models require longer and more complex prompts to yield impressive outcomes, which is less user-friendly. Thus, we provide a powerful and user-friendly pipeline to generate more elaborate images through iterative refinement with the aid of large language models.
\subsection{Synthetic Data for Visual Tasks}
Recently, there has been an increasing interest in using high-quality synthetic data to augment training data for downstream tasks~\cite{lin2023explore, he2022synthetic, benigmim2023one, bansal2023leaving}. PET~\cite{schick2020s} primarily focuses on a semi-supervised situation to automatically generate abundant labeled data for augmentation. \cite{he2022synthetic} use GLIDE to generate abundant class-conditioned images and explore the effectiveness of synthetic data for image recognition tasks in data-scarce settings. For the task of few-shot object detection, a method proposed in~\cite{lin2023explore} involves selecting representative samples from a large-scale synthetic dataset to potentially enhance the performance of FSOD models. Here we present a novel approach for generating high-quality synthetic data by leveraging state-of-the-art text-to-image models and LLMs. Our method eliminates the need for expensive prompt engineering by introducing a unified framework that produces abundant and elaborate images with annotations in a single pipeline.
\section{Method}
ChatGenImage is a labeling collaboration framework that involves a mentor large language model~(LLM) and numerous AIGC models as controllable image creators, and labeling foundation models for executing the labeling task. The workflow of Label Anything consists of two stages: \textbf{Language Enhancement Image Initialization} and \textbf{Iteratively Local Refinement and Labeling}, as shown in Figure~\ref{architecturel}. 1) During the first stage, An LLM (\textit{e.g., ChatGPT}) analyze the label word from the user input, and generate complex scene descriptions and object-centric descriptions in \textit{Global Prompts Brainstorming}. Then, AIGC generators initialize controllable images based on the global constraints from the LLM. 2) During the second stage, the LLM produces \textit{Label Editing Prompts} based on the high-quality pseudo labels automatically obtained from the \textit{Label Foundation Toolkit} and employs them to iteratively control the process of local image editing. Based on that, AIGC models can perform \textit{Controllable Image Editing} from both background and foreground to obtain more diversified synthetic images. Through our approach, the generated images are modified to align with the complex annotations, resulting in high-quality synthetic data suitable for data augmentation in downstream tasks.
\subsection{Global Prompts Brainstorming} Due to the limited knowledge of the large language model about the AIGC model, it is not capable of providing appropriate prompts for the AIGC model. Therefore, ChatGenImage utilizes a hybrid approach that combines both specification-based instruction and demonstration-based learning to effectively guide the AIGC model in generating high-quality synthetic data.

\noindent\textbf{Specification-based Instruction.} The prompt specification can serve as a standard template for large language models to comprehend visual attributes of the specific concept, thereby facilitating the sensible scene imagination for a given word through slot filling. However, using category names alone in AIGC models may limit their ability to perceive visual features, leading to ambiguous image generation\cite{he2022synthetic}. To help the large language model imagine effective scene descriptions, ChatGenImage prompts focus on descriptive features rather than broad categories. In the first stage of ChatGenImage, the large language model take the \textit{Label Word} from the user and construct several relevant descriptions as its \textit{Visual Feature} for global prompt brainstorming. Moreover, we propose to automatically obtain appropriate visual features by prompting the LLM to describe the visual features that distinguish that category in a photograph. We demonstrate the prompt process for visual feature descriptions and controllable generation in Table \ref{prompt_example}.

\noindent\textbf{Demonstration-based learning.} ChatGenImage utilizes the in-context learning capability of LLMs and injects several demonstrations into the prompt learning, helping large language models to better understand the parameter criteria for conditional image generation. Each demonstration is a group of input and output on scene prompts brainstorming——the user's request in standard templates and the expected image descriptions for AIGC models. Furthermore, these demonstrations consist of complex scene descriptions and object-centric descriptions, as shown in Figure \ref{architecturel}, effectively aid ChatGenImage in understanding the given label's attributes in various environments and imagining reasonable prompts for high-quality image synthesis.


\begin{table*}[ht]
\centering
\begin{tabular}{|m{0.4cm}|m{3.5cm}m{12cm}|}
\hline
 \multirow{6}{*}{\rotatebox{90}{Visual Descriptor}} & \multicolumn{2}{c|}{Prompt} \\ [2pt]  \cline{2-3}
   & \multicolumn{2}{L{15.85cm}|}{"role": "system", "content": "You are an expert in the field of vision and graphics, please fully consider the input concept or topic, give me the most important fine-grained visual features of the input concept or category based on the Wikipedia. Only give me several phrases or keywords as more as possible."

  "role": "user", "content": "Q: What are useful visual features for distinguishing a \{\textbf{category name}\} in a photo?
  
  A: There are several useful visual features to tell there is a \{\textbf{category name}\} in a photo:"}\\\hline 
   \multirow{23}{*}{\rotatebox{90}{AIGC Creator}} & \multicolumn{2}{c|}{Prompt}  \\\cline{2-3}
   &  \multicolumn{2}{L{15.85cm}|}{"role": "system",
        "content": "The AI assistant is a professional data specialist. The AI assistant can imagine different real scenes according to the input caption and return the background and closely related foreground objects in a specific format. Note that the background should be described with a single word or phrase, and each background contains five main objects." \{\textbf{Demonstration}\}

    "role": "user",
  "content": "Give me \{\textbf{prompt number}\} high quality prompts for text-to-image models about the amazing close-up realistic scene containing the \{\textbf{label word}\}. Scene prompts should also contain other objects as more as possible. The \{\textbf{label word}\} is with \{\textbf{visual feature}\}."}\\\cline{2-3}
  & \multicolumn{2}{c}{Demonstration} \vline \\\cline{2-3}
  & \multicolumn{2}{L{15.85cm}|}{Sure, here are 1 high quality prompts for text-to-image models about the amazing close-up realistic scene containing the red panda: Bring a photo-realistic close-up scene of a red panda to life, set against a backdrop of a rustic temple in a mountainous region. The panda should be situated in front of the temple, with its round ears, short legs, and bushy tail with alternating red and white rings visible. The temple should have ornate carvings and architecture, with stone lanterns and a small pond nearby. The panda's white paws with black claws should be visible as it sits atop a stone pedestal, perhaps with some cherry blossom petals scattered around it. The overall scene should have a peaceful and serene atmosphere.} \\ \cline{2-3}
    & \multicolumn{2}{c}{Visual Feature} \vline \\\cline{2-3}
     & \centering{Label word: "Lemur"} & 
    - Large, forward-facing eyes with binocular vision; Soft, thick fur covering the body
    
    - Long, bushy tail; Striking coloration patterns such as black and white rings
    
    - Slender fingers and toes with opposable thumbs for grasping and climbing;  Prominent ears that can be pointed or rounded with tufts of fur
    
    - Wet, reflective nose; Rounded head with a shortened snout and large ears
    
    - Relatively small body size, typically weighing between 2 and 5 kilograms; Distinctive vocalization or call that can vary between species and subspecies \\\hline
    \multirow{9}{*}{\rotatebox{90}{Scene Imagination}} & \multicolumn{2}{c|}{Prompt}  \\\cline{2-3}
   &  \multicolumn{2}{L{15.85cm}|}{"role": "system", "content": "The AI assistant is a professional data specialist. The AI assistant can imagine different real scenes according to the input caption and return the background and closely related foreground objects in a specific format. Note that the background should be described with a single word or phrase, and each background contains five main objects."

   "role": "user", "content": "Give me {scene number} real scene descriptions based on the context {caption}. The scene objects should consist {exist objects}, and also contain five additional objects associated with the background. Each scene description is a complex sentence containing the above objects. Return the result in the following format: {'background':[], 'objects':[], 'description':[]}. Only return the result."}\\\hline
       \multirow{12}{*}{\rotatebox{90}{Box Candidates Generation}} & \multicolumn{2}{c|}{Prompt}  \\\cline{2-3}
   &  \multicolumn{2}{L{15.85cm}|}{"Please make \{\textbf{prompt number}\} possible prediction of the remaining box coordinates with different box size based on the dense description "\{\textbf{caption}\}". Note that the image size is (512,512), and the existing box coordinates are [\{\textbf{existing boxes info}\}]. Based on the layout of the objects, predict the possible number reasonable box coordinates of the following objects \{\textbf{target objects}\}. The size of the \{\textbf{target objects}\} box should be based on the category and the size of other object boxes, and the width and height of the box should be greater than 75 and less than 300. Only return each result in the following format: {"\textcolor{red}{label}":, "\textcolor{red}{box}":, "\textcolor{red}{relationship}":}"}\\\cline{2-3} 
   & \multicolumn{2}{c|}{Demonstration}  \\\cline{2-3}
   &  \multicolumn{2}{L{15.85cm}|}{"\{\textbf{caption}\}": 'there is a dog sitting on a bench in a field.'
   
   "\{\textbf{existing boxes info}\}": {"value": 1, "label": "bench", "logit": 0.84, "box": [33.93, 224.34, 463.20, 491.01]}, {"value": 2, "label": "dog", "logit": 0.43, "box": [175.71, 116.29, 311.58, 367.13]}
   
   Return Results: {"\textcolor{red}{label}": '\textcolor{blue}{cat}', "\textcolor{red}{box}": \textcolor{blue}{[343.23, 176.29, 467.23, 353.13]}, "\textcolor{red}{relationship}": '\textcolor{blue}{sitting next to the dog.}'}}\\\cline{2-3} 
    
\hline
\end{tabular}
\caption{The details of the prompt design in ChatGenImage. There are injectable slots in the prompts, such as \textbf{Caption}, \textbf{Visual Feature}, and \textbf{Existing Boxes Info}. These slots imply visual perceptions that help LLM building multimodal awareness and are uniformly replaced with the contents from visual foundation models before being fed into the LLM.}
\label{prompt_example}
\end{table*}
\subsection{Label Foundation Toolkit}
Since ChatGPT is a pure language model and cannot ``see” any visual information, we present the initialized images to several powerful label foundation toolkits~(\textit{i.e.}, Segment Anything Model~\cite{kirillov2023segment}, Grounding DINO~\cite{liu2023grounding}, and BLIP2~\cite{li2023blip}) and serve them as sensors in the system to provide perceptual information to the ChatGPT.

\noindent\textbf{Segment Anything Model~(SAM)} is a large ViT-based model trained on the large visual corpus~(SA-1B)~\cite{kirillov2023segment}, which has demonstrated promising zero-shot segmentation capabilities in various scenarios and the great potential for data labeling in visual tasks. But it needs precise prompts~(like boxes/points) to generate accurate masks and lacks category predictions or annotations for each mask.

\noindent\textbf{Grounding DINO} is a strong zero-shot detector which is capable of to generate high quality boxes and labels with free-form text~\cite{liu2023grounding}, which can also serves as box prompts generator for SAM. Our approach combines the strengths of Grounding DINO and SAM to detect and segment comprehensive regions in each synthetic image. This builds a powerful pipeline for complex visual scene labeling and produces abundant fine-grained pseudo labels for training.

\noindent\textbf{BLIP2} is a language-vision model that seamlessly integrates visual input into text sequences to facilitate overall visual perception of LLMs~\cite{li2023blip}. By combining BLIP2 with the aforementioned visual models, our approach can automatically generate high-quality text descriptions for synthetic images. The LLMs then use these descriptions to understand image regions and return local editing prompts for controllable and diverse image refinement.
\subsection{Local Editing Prompt}

Despite the ability of AIGC models to generate labeled images through prompt-based techniques with the aid of LLMs, their effectiveness in depicting intricate scenes and fine-grained attributes remains limited due to their inclination to create object-centric images with only global constraints. Besides, we observe that the generated images contain fewer objects, which poses a challenge in constructing complex scenes for demanding downstream tasks~(\textit{e.g.}, scene graph generation and visual question answering). Thus we further introduce local editing prompt in the iterative pipeline for fine-grained image refinement.

In detail, we design a iterative communication that encourages ChatGPT to provide a series of informative feedbacks based on the generated images from AIGC models and corresponding labels from label foundation toolkits. Since language models are blind to the initialized image, ChatGPT cannot partially edited the initial image directly. Hence, we employ a predefined prompt template and populate the slots with the corresponding caption and object box coordinates identified by the visual foundation models. This template is subsequently utilized by the LLMs to produce novel scenes that comprise new backgrounds and additional objects. It is worth noting that ChatGPT can voluntarily select a reasonable location for editing based on human instructions and its underlying knowledge, autonomously generating accurate local editing prompts. Then the AIGC model use the resulting prompts to edit the images and improve their quality by adding missing details.

\begin{algorithm}
	\renewcommand{\algorithmicrequire}{\textbf{Input:}}
	\renewcommand{\algorithmicensure}{\textbf{Output:}}
	\caption{ChatGenImage Pipeline}
	\label{alg1}
	\begin{algorithmic}[1]
		\STATE Initialization: Label $w_0$, Init Image $I_0={\rm SD}(w_0)$
		\STATE  Caption $c={\rm BLIP}(I_0)$
            \STATE Object Boxes $b={\rm GroundingDINO}(I_0)$
            \STATE Segment Masks $m={\rm SAM}(I_0)$, iterative $i$
            \STATE ${\rm Image} = \{(I_0,c,b,m)\}$
		\REPEAT
		\STATE $i \leftarrow i - 1$
            \STATE Scenes = ${\rm ChatGPT}(c, b, m)$,  
            $n \leftarrow $length(Scenes)
            \FOR{$i=0$ to $n$}
            \STATE Scene=Scenes[$i$]
            \STATE Image $I$= Editing$(I_0, {\rm Scene}['background'])$
            \STATE Update $I$= Filling$(I_0, {\rm Scene}['objects'])$
		\STATE Update $c, b, m$ by `Label Foundation Toolkit'
            \STATE Update Image = ${\rm Image} \cup \{(I,c,b,m)\}$ if not Filter($I$)
            \ENDFOR
		\UNTIL $i == 0$
		\ENSURE ${\rm Image} = \{(I_i,c_i,b_i,m_i)\}$
	\end{algorithmic}  
\end{algorithm}
\subsection{Controllable Image Editing}
To efficiently generate a significant amount of images with complex scenes and rich annotations in a low-resource manner, It is necessary to collaborate with various controllable editing models that can perform controllable image editing based on both global and local prompts. The total process is shown in Algorithm~\ref{alg1}. Besides, we use image filtering rules to figure out those representative samples as valid results and utilize the label foundation toolkit to get high-quality annotations for downstream tasks.

\noindent\textbf{Background Imagination.}
We notice that retrieved or directly generated images are usually restricted to to a single domain, thereby leading to a constraint in the development of robust visual models~\cite{dunlap2022using}. Furthermore, it is impractical to obtain labeled data for all possible anticipated settings at once is often impractical due to the significant expense involved. However, acquiring linguistic knowledge of the anticipated domain shift is a more cost-effective and accessible approach. 

Hence, we leverage the ChatGPT to generate novel backgrounds for the original image and employ InstructPix2Pix~\cite{brooks2022instructpix2pix} to substitute different backgrounds, generating a vast collection of composite images across various domains in a cost-effective manner. To preserve the foreground semantics of images after background modification, we perform target detection on both the original and modified images, and apply filter rules to exclude images with missing objects.

\noindent\textbf{Foreground Object Filling.}
To avoid altering the semantic information of the source image, we propose a method to increase the complexity of the image scene by filling foreground objects. The necessary object labels, coordinates, and their interactions with the scene~(\textit{i.e.}, \{label: `rocks', box: [200, 50, 300, 150], relationship: `near the cabin and surrounded by trees'\}) can be obtained by filtering the local editing prompts automatically. Once collect sufficient possible boxes through ChatGPT, we can use Blended Latent Diffusion~\cite{avrahami2022blended,avrahami2022blended2} to fill novel objects in the specific position of the foreground. It is worth noting that we filter out the overlapping bounding boxes to ensure that the original semantics are preserved. In this way, we greatly enrich the spatial interaction of objects in the generated image.
\begin{figure}[ht]

\centering
   \includegraphics[width=1.0\linewidth]{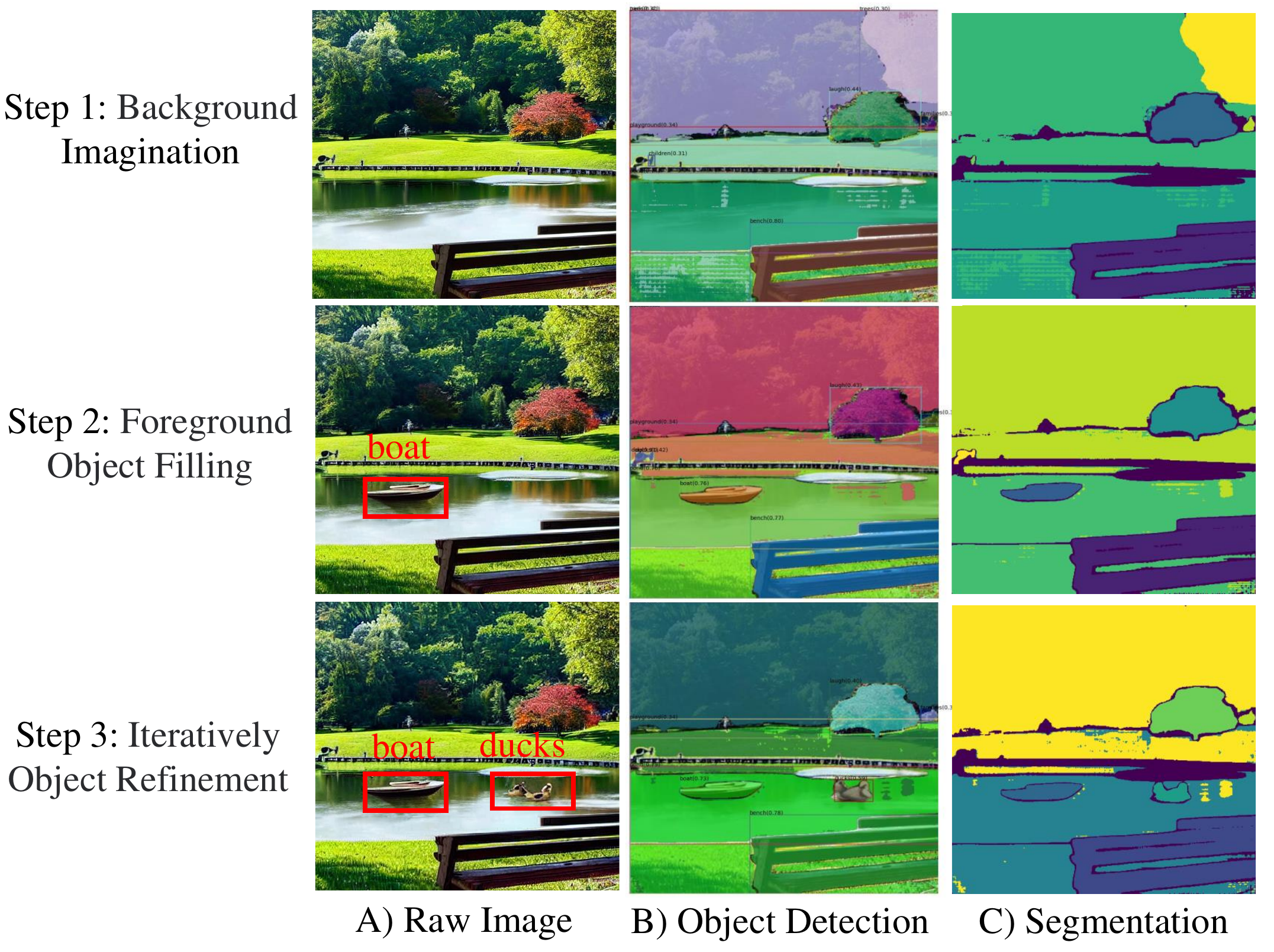}
    
   \caption{Visualization results of Iteratively Local Refinement and Labeling. It contains three steps: 1) Background Imagination, 2) Iteratively Object Filling, 3) Label anything in the image via visual foundation models.}
   \label{example}
\end{figure} 
\begin{figure*}[ht]

\centering
   \includegraphics[width=1.0\linewidth]{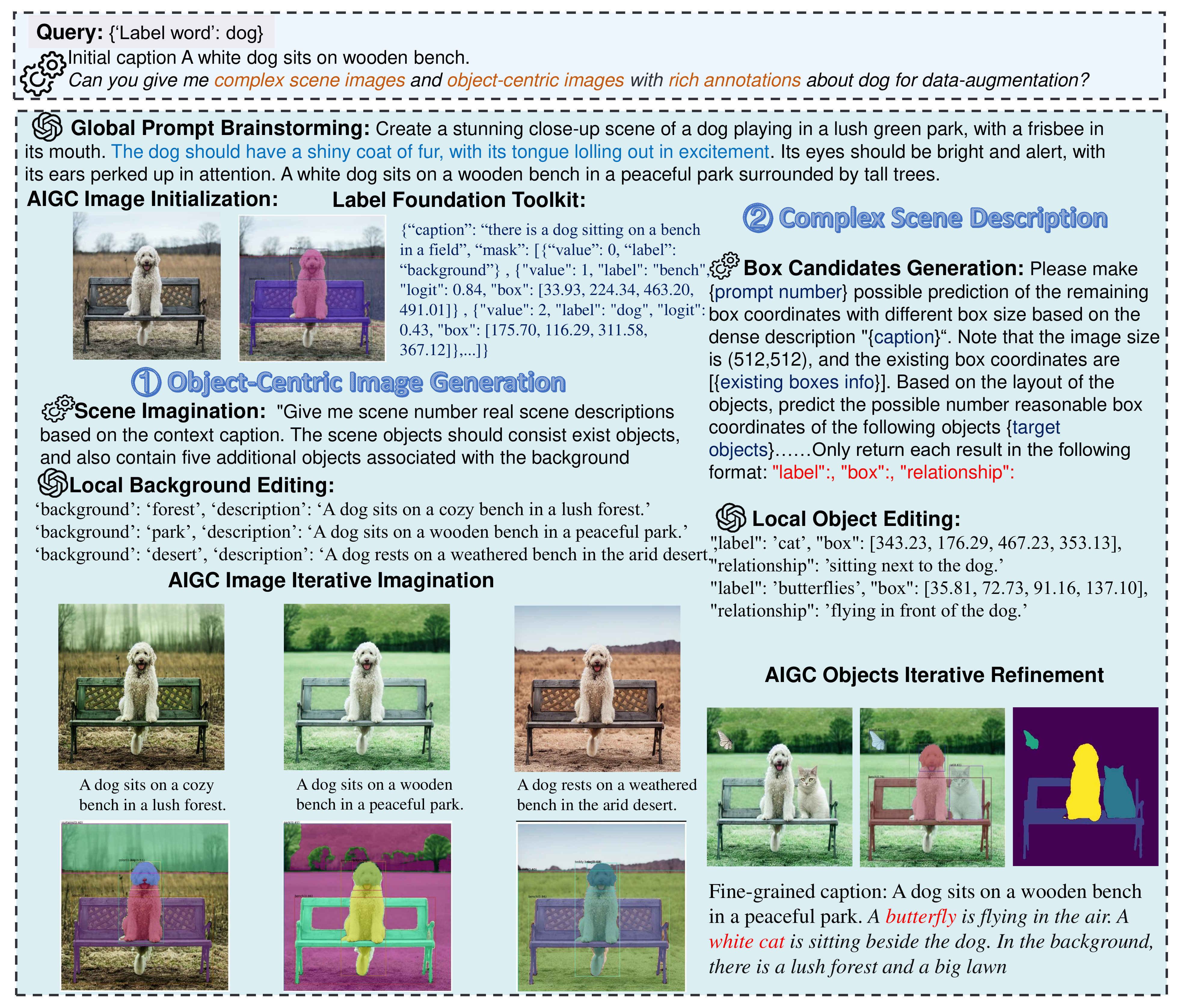}
    
   \caption{Qualitative analysis of object-centric image generation and complex scene description with multiple background and objects.}
   \label{object_pipeline}
\end{figure*}
\begin{figure*}[ht]

\centering
   \includegraphics[width=1.0\linewidth]{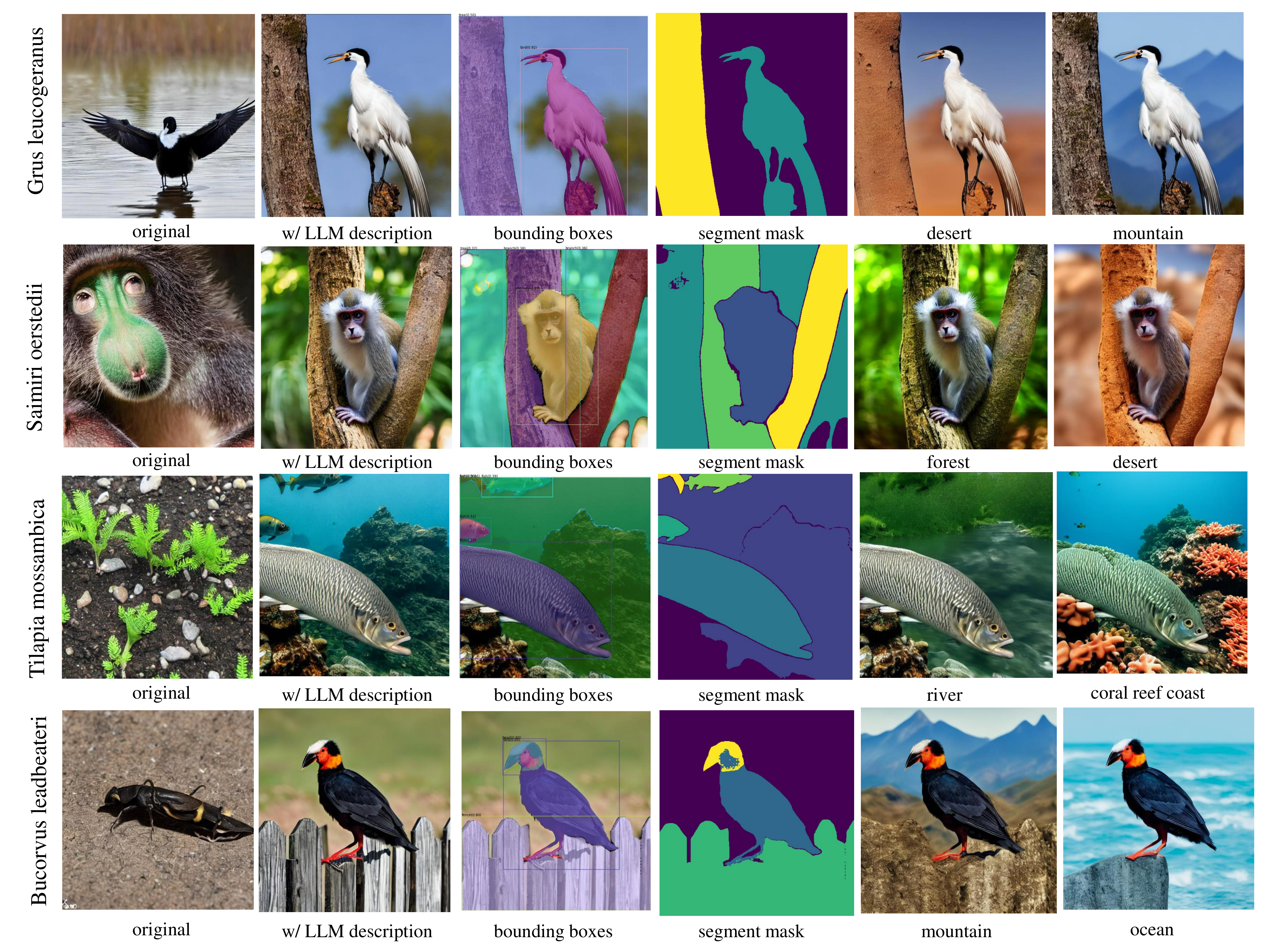}
    
   \caption{Visualization results of ChatGenImage for object-centric image generation.}
   \label{example_objects}
\end{figure*}
\subsection{Image Filtering Rules}
Though Most state-of-the-art AIGC methods generate astonishing images, they may have several visual and textual issues: 1) boundary incoherence, 2) semantic mismatch and 3) object missing. It is essential to establish robust image filtering rules that can effectively evaluate the synthetic images and filter out those low-quality results. To address above challenges, we introduce a Pixel Checking~(\textbf{PC}) and a Semantic Checking~(\textbf{SC}) strategy for the generated images from the perspective of visual pixels and textual semantics. 

\noindent\textbf{Pixel Checking.} To ensure the boundary consistency of the edited image, we evaluate the fidelity of the generated image. IS~\cite{salimans2016improved}, FID~\cite{heusel2017gans} SceneFID\cite{sylvain2021object} are common metrics to evaluate the fidelity of general images in different scales. 
However, all of these metrics rely on ground truth labels, which are not suitable for assessing images generated by stable diffusion models~\cite{ramesh2022hierarchical}.
Therefore, we exploit the SSIM and PSNR~\cite{hore2010image} to explore structural similarity and pixel similarity between the locally edited image and the original image for pixel checking. We employ a threshold strategy of PSNR and SSIM between the original and edited images, minimizing artifacts and incoherence at the editing boundary to preserve the global coherence of the image during the local editing process.

\noindent\textbf{Semantic Checking.} 
Considering that local image editing may introduce undesired items to destroy the semantics of the original image, we evaluate the semantics and object detection of the generated image during semantic checking.
Specifically, we generate a set of image candidates based on scene descriptions of specific label words during both global initialization and local image editing. Then, we use the CLIP similarity score~\cite{radford2021learning} to evaluate the semantic alignment between the image candidates and textual constraints. We rank the image candidates based on the score and filter out the low-confidence images to obtain most matching ones as the final result. Besides, as the we employ open vocabulary object detection on the images after background editing. We only retain those images that can identify the novel background and original foreground objects to keep the original semantics and enhance the downstream utility of the edited images.

\begin{figure*}[ht]

\centering
   \includegraphics[width=1.0\linewidth]{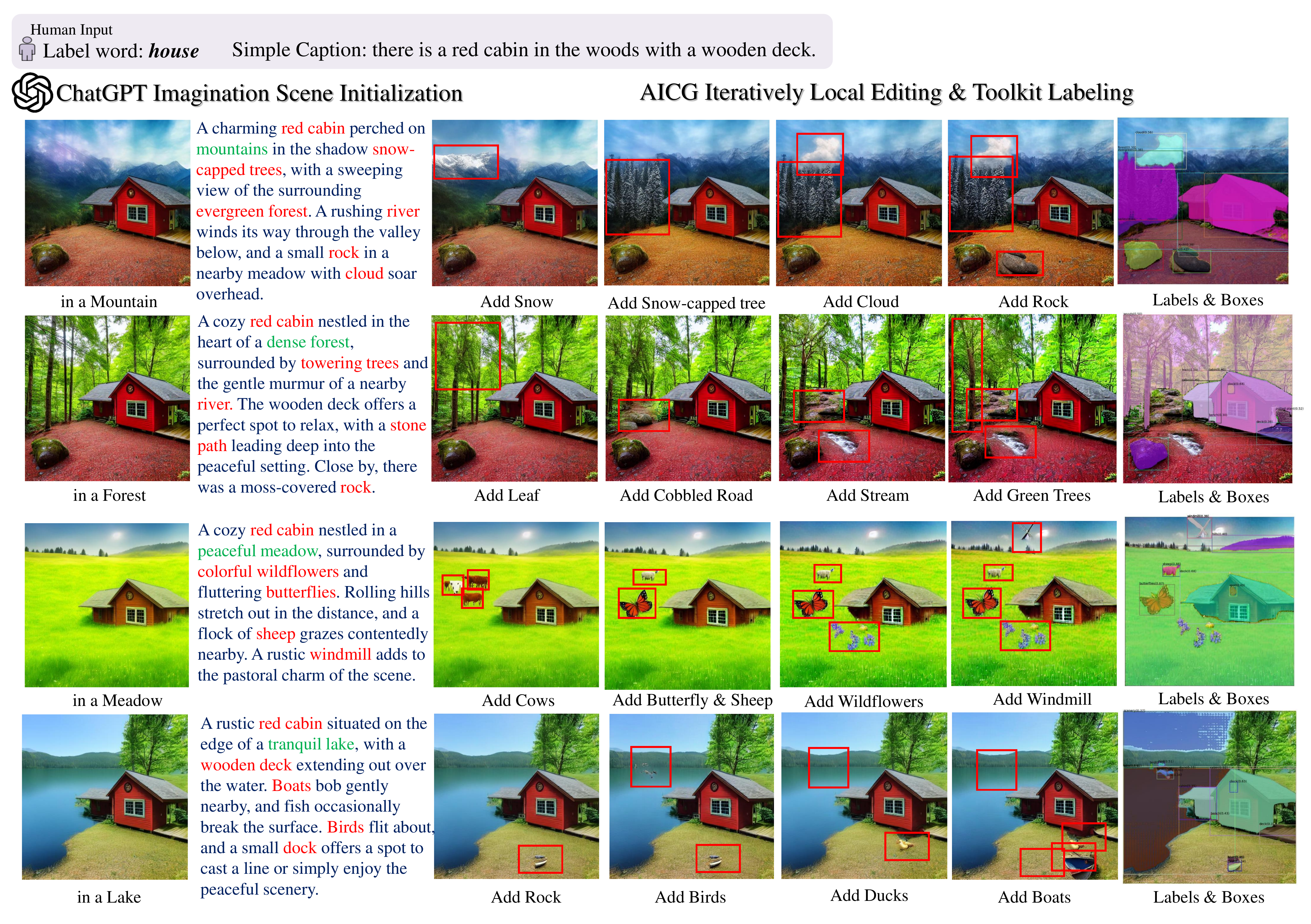}
    
   \caption{Visualization results of ChatGenImage for complex scene imagination.}
   \label{example_scene}
\end{figure*}

\section{Experiments}
In this version, we discuss our main experiments setup and several results. We demonstrate the effectiveness of our ChatGenImage on interactive data synthesis through qualitative results, showing the potential of using ChatGenImage for systematic vision adaptation. In the next release, we will further explore how to better leverage the synthetic data obtained from our ChatGenImage framework for better downstream task generalization.
\subsection{Setting}In our experiments, we employed the gpt-3.5-turbo variants of the GPT models as the large language models~(\textit{i.e.}, ChatGPT), which are publicly accessible through the OpenAI API\footnote{\url{https://platform.openai.com/}}. To make the LLM output more stable, we set the decoding temperature to 0. For AIGC models, we uniformly set the pixel of the picture to 512×512 to save the memory overhead. Also to adapt to the whole system, we use stable diffusion v1.5 as the AIGC base model with the same default parameters as the original setting~\cite{rombach2022high}. We provide detailed prompts designed for the Visual Descriptor, AIGC Creator, Scene Imagination, Box Candidates Generation in the step of \textit{Global Prompts Brainstorming} and \textit{Local Editing Prompts} in Table~\ref{prompt_example}, where \textbf{\{variable\}} indicates that the slot needs to be populated with the corresponding text before the prompt can be fed into the LLM. This label pipeline is based on a single Nvidia RTX 3090 GPU, which is affordable for most people.

\subsection{Qualitative Results}
We evaluate the generated image and fine-grained annotations in our ChatGenImage in two cases~(\textit{i.e.}, complex scene description and object-centric image generation). In Figures~\ref{object_pipeline}, we show several dialogue demonstrations and qualitative analysis for above two different cases of requirement data, respectively. We collect several label words of rare and endangered species for testing, which have few photos in the web and are unfamiliar to most image classifiers. We compare our approach with LLM descriptions and original images generated by naive AIGC models. The result is shown in Figure~\ref{example_objects}. The experimental result indicates that the proposed ChatGenImage is both general and controllable, effectively creating robust images even for unfamiliar and rare concepts. Through interactive communication with the LLM, the AIGC can learn the specific descriptions of novel concepts and complete controllable generation in different domains via iterative image editing.

Besides, we explore the effectiveness of the ChatGenImage for complex scene descriptions and whether the LLM help AIGC models iteratively fill accurate objects in the foreground.
In Figure~\ref{example_scene}, we show that the LLM provides several information comprising of object coordinates and relation interactions, which is then used by Stable Diffusion to generate diverse backgrounds~(\textit{e.g.}, mountain, forest) and incorporate relevant objects~(\textit{e.g.}, snow trees, stream and rocks) to produce a rich image depicting a complex scene.

\subsection{Applications}
ChatGenImage, as an interactive data synthesis framework, provides two generation modes and fine-grained images with various domains to easily meet the requirements of different field. Moreover, the usage of synthetic data from ChatGenImage enables more generalizable of downstream tasks in complex application scenarios.

Since ChatGenImage can iteratively generate large amount of diverse images via LLM-AIGC collaboration, it can provide extra unseen data domains for systematic vision adapation. We will evaluate ChatGenImgae on several domain adaptation benchmarks to investigate how to enrich the existing dataset with synthetic data and construct adaptive visual perception system in a cost-less manner.

\section{Conclusion}
ChatGenImage is a versatile tool that combines the capabilities of LLMs, AIGC models and powerful label foundation toolkits. Based on the collaboration of diverse models, ChatGenImage enables to generate fine-grained images with rich annotations for data augmentation. In the future, we will further develop our approach to support more complex and challengeable scenarios, like fine-grained human-object interaction, action editing, etc., and apply it to more realistic applications for systematic vision adaptation.

{\small
\bibliographystyle{ieee_fullname}
\bibliography{egbib}
}

\end{document}